\documentclass[letterpaper]{article} % DO NOT CHANGE THIS
\usepackage{aaai23}  % DO NOT CHANGE THIS
\usepackage{times}  % DO NOT CHANGE THIS
\usepackage{helvet}  % DO NOT CHANGE THIS
\usepackage{courier}  % DO NOT CHANGE THIS
\usepackage[hyphens]{url}  % DO NOT CHANGE THIS
\usepackage{graphicx} % DO NOT CHANGE THIS
\usepackage{natbib}  % DO NOT CHANGE THIS AND DO NOT ADD ANY OPTIONS TO IT
\usepackage{caption} % DO NOT CHANGE THIS AND DO NOT ADD ANY OPTIONS TO IT
\usepackage{algorithm}
\usepackage{algorithmic}
\usepackage{amsfonts}
\usepackage{amsmath}
\usepackage{multirow}

\usepackage{newfloat}
\usepackage{listings}
\DeclareCaptionStyle{ruled}{labelfont=normalfont,labelsep=colon,strut=off} % DO NOT CHANGE THIS
\floatstyle{ruled}
\newfloat{listing}{tb}{lst}{}
\floatname{listing}{Listing}
\title{Rich Event Modeling for Script Event Prediction}
\author{
    Long Bai,
    Saiping Guan,
    Zixuan Li,
    Jiafeng Guo,
    Xiaolong Jin,
    Xueqi Cheng
}
\affiliations{
    CAS Key Lab of Network Data Science and Technology, \\
    Institute of Computing Technology, Chinese Academy of Sciences (CAS); \\
    School of Computer Science and Technology, University of Chinese Academy of Sciences \\
    \texttt{\{bailong18b,guansaiping,lizixuan,guojiafeng,jinxiaolong,cxq\}@ict.ac.cn}
}

\usepackage{bibentry}
\begin{document}

\maketitle

\begin{abstract}
	% 1. Basic definition of script and script event prediction.
	Script is a kind of structured knowledge extracted from texts, which contains a sequence of events.
	Based on such knowledge, script event prediction aims to predict the subsequent event.
	% 2. The key challenge: event representation method.
	To do so,
	two aspects should be considered for events,
	namely, event description (i.e., what the events should contain) and event encoding (i.e., how they should be encoded).
	% 3. The problem of existing approaches.
	Most existing methods describe an event by a verb together with only a few core arguments 
	(i.e., subject, object, and indirect object),
	which are not precise.
	In addition, existing event encoders are limited to a fixed number of arguments,
	which are not flexible to deal with extra information.
	% 4. The proposed event representation method.
	Thus, in this paper, we propose the Rich Event Prediction (REP) framework for script event prediction.
	Fundamentally, it is based on the proposed rich event description,
	which enriches the existing ones with three kinds of important information,
	namely, the senses of verbs, extra semantic roles, and types of participants.
	REP contains an event extractor to extract such information from texts.
	% 5. How to encode the events.
	Based on the extracted rich information, a predictor then selects the most probable subsequent event. 
	The core component of the predictor is a transformer-based event encoder to flexibly deal with an arbitrary number of arguments.
	% 6. How does the proposed method perform.
	Experimental results on the widely used Gigaword Corpus show the effectiveness of the proposed framework.
	
\end{abstract}

\section{Introduction}
% 1. Introduction of script and script event prediction
Script is a typical kind of knowledge to describe daily scenarios~\cite{abelson1977scripts},
which is usually in the form of event sequence.
Recently, a kind of scripts extracted from texts, called narrative event chain~\cite{chambers-jurafsky-2008-unsupervised}, has attracted much attention,
where events sharing a common participant (called protagonist) are temporally ordered into a sequence.
Script Event Prediction (SEP) task aims to predict the subsequent event based on the historical narrative event chain.
It is helpful for a number of natural language processing tasks,
such as coreference resolution~\cite{bean-riloff-2004-unsupervised},
discourse understanding~\cite{lee-goldwasser-2019-multi}, and story generation~\cite{chaturvedi-etal-2017-story}.

% 2. Introduction of the key challenges.
In this task, the essential element, i.e., the event, 
consists of a verb and multiple arguments.
Such a complex structure brings challenges to the SEP task.
Specifically, the challenges come from two aspects, 
namely, event description and event encoding.
The former is about what the events should contain,
while the latter concerns how they should be encoded into machine-computable representations.

\begin{figure}
	\centering
	\includegraphics[scale=0.30]{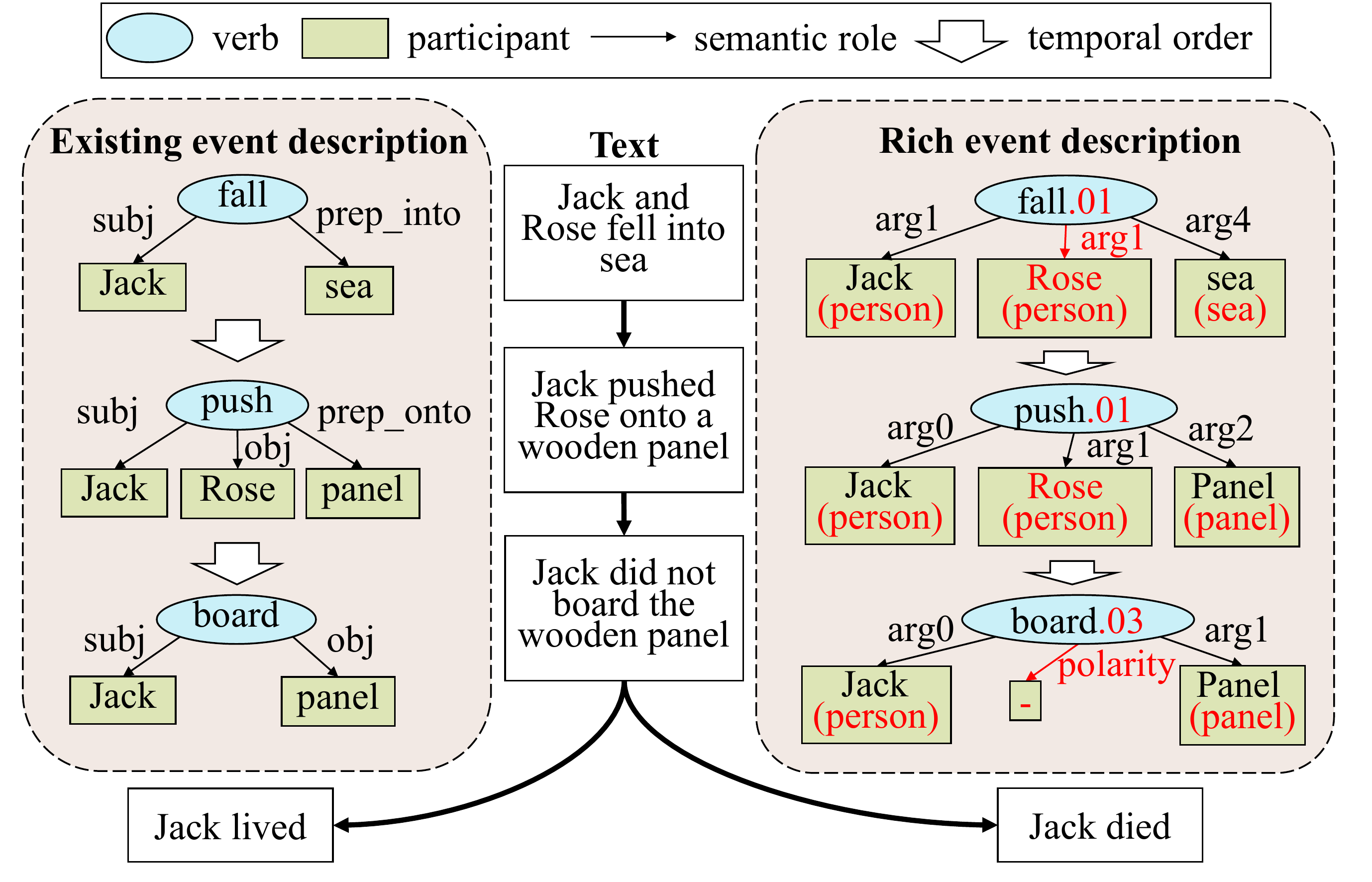}
	\caption{An example of events under existing event description and rich event description. 
		The enriched information is highlighted in red.
		``Jack lived'' is more likely to happen under existing event description,
		while ``Jack died'' is more likely to happen under rich event description.
		From the texts, ``Jack died'' is more likey to happen.}
	\label{fig:introduction}
\end{figure}

% 3. Limitations of current event definition
With respect to event description, in the existing methods, 
each event is represented by a verb and three arguments.
Each argument contains a semantic role (i.e., subject, object, or indirect object)
~\footnote{Precisely, they are grammar roles. 
	In this paper, we use the word ``semantic role'' for convenience.} 
and the corresponding participant~\cite{Granroth-Wilding_Clark_2016}.
Each participant is usually represented by the most salient headword of coreferred participant mentions
to induce the generalized semantic knowledge among mentions
~\cite{chambers-jurafsky-2009-unsupervised}.
However, such event description faces three limitations.
% Problem of verbs
(1) The verb is ambiguous, which leads to a misunderstanding of the event.
For example, in Figure~\ref{fig:introduction}, 
the verb ``fall'' can be explained as ``move downward'' or ``be defeated''.
This event description is hard to distinguish different meanings.
% Problem of role types
(2) In some situations, the current three participants are insufficient to precisely describe an event.
Existing methods are not able to relate a role type to multiple participants.
For example, the event ``Jack and Rose fell into sea'' contains a compound subject, ``Jack and Rose'',
which includes two participants.
In addition, the modifiers of the events are not carefully considered.
For example, ``not'' in ``Jack did not board the panel''
leads the occurrence of the event to the opposite.
% Problem of head words
(3) Headwords are not able to well describe the type information of participants.
Though, in some cases, existing methods may successfully obtain the types via headwords, 
e.g., ``a wooden panel'' is a panel,
they fail to know that ``Jack'' and ``Rose'' are both humans.
% Conclusion
All these limitations bring challenges to precisely describing an event.

% 4. Limitations of current event encoding
With respect to event encoding, 
most existing methods usually apply a Multi-Layer Perceptron (MLP) network to encode events into low-dimensional vectors~\cite{Granroth-Wilding_Clark_2016},
which only supports a fixed number of arguments.
In order to capture more subtle semantic interactions between a verb and its arguments,
some methods first encode verb-argument pairs and then aggregate them~\cite{Weber_Balasubramanian_Chambers_2018,ding-etal-2019-event-representation}.
However, it is still challenging to flexibly integrate an arbitrary number of arguments 
while obtaining sufficient interactions among the verb and the arguments.

% 5. Detailed implementation
To deal with the above challenges, 
in this paper, we propose the Rich Event Prediction (REP) framework,
based on the proposed rich event description.
Compared with the existing event description,
rich event description contains the senses of verbs, extra semantic roles and types of participants,
which are able to express the event more precisely.
The REP framework predicts the subsequent event via two main modules, namely, the event extractor and predictor.
The event extractor extracts rich event information from texts via an intermediate semantic representation, 
i.e., Abstract Meaning Representation (AMR)~\cite{banarescu-etal-2013-abstract}.
With the rich events as inputs, the predictor further projects them into low-dimensional vectors via a rich event encoder,
and then predicts the most probable subsequent event.
The rich event encoder utilizes a transformer-based network to capture the subtle interactions among the verb and the arbitrary number of arguments.
% 6. Conclusion
In general, the main contributions of this paper are as follows:
\begin{itemize}
	\item We propose the rich event description to precisely express the events,
	which additionally contains three kinds of important information, 
	namely, senses of verbs, extra semantic roles, and types of participants.
	
	\item We propose a predictor for script event prediction which flexibly capture the subtle interactions among the verb and the arbitrary number of arguments via the designed rich event encoder.
	
	\item We conduct extensive experiments on the widely used Gigaword corpus,
	which show the superiority of the proposed framework.
	
\end{itemize}

\section{Related Work}
SEP is to predict the subsequent event of a given narrative event chain~\cite{chambers-jurafsky-2008-unsupervised}.
Events in the chain share a common entity (called the protagonist) and are ordered by their temporal relations.
The research line on SEP starts from \cite{chambers-jurafsky-2008-unsupervised}.
It describes an event by a tuple $\langle verb, dependency \rangle$, 
which denotes the verb and its dependency relation with the protagonist. 
Then, to model the relevance among events,
it applies Pointwise Mutual Information (PMI) to get the score of each event pair.
Finally, these pairwise scores are aggregated to predict the subsequent event of the given narrative event chain.

The following studies focus on handling the two essential problems of SEP,
namely, event modeling and relevance modeling.
In this paper, we mainly focus on event modeling, 
which consists of event description (i.e., what the events should contain) and event encoding (i.e., how they should be encoded).

With respect to the event description,
\citet{balasubramanian-etal-2013-generating} propose to use $\langle subj, verb, obj \rangle$ triple to capture the co-occurrence between the subject and object.
\citet{pichotta-mooney-2014-statistical,Granroth-Wilding_Clark_2016} additionally take the indirect object into consideration.
Currently, most studies on SEP are based on this $\langle verb, subj, obj, iobj \rangle$ description
and use headword to represent each participant following \citet{chambers-jurafsky-2009-unsupervised}.
\citet{Lee_Goldwasser_2018} additionally consider the sentiments of events and the animacies of arguments.
They also consider the negations of the verb,
but they directly turn it into another verb, such as ``eat'' and ``not\_eat''.
However, they only consider this one kind of modifier and fail to model the others.
Moreover, the relevance between a verb and its negation is difficult to be captured.
The difference between our event description and the previous ones is that,
ours is more flexible to handle an arbitrary number of arguments and different kinds of modifiers,
which is closer to the nature of events.
Since this paper mainly discusses what a structured event contains,
we do not consider to directly describe an event by its original text, like~\cite{lee-etal-2020-weakly,bai-etal-2021-integrating}.

\begin{figure*}
	\centering
	\includegraphics[scale=0.40]{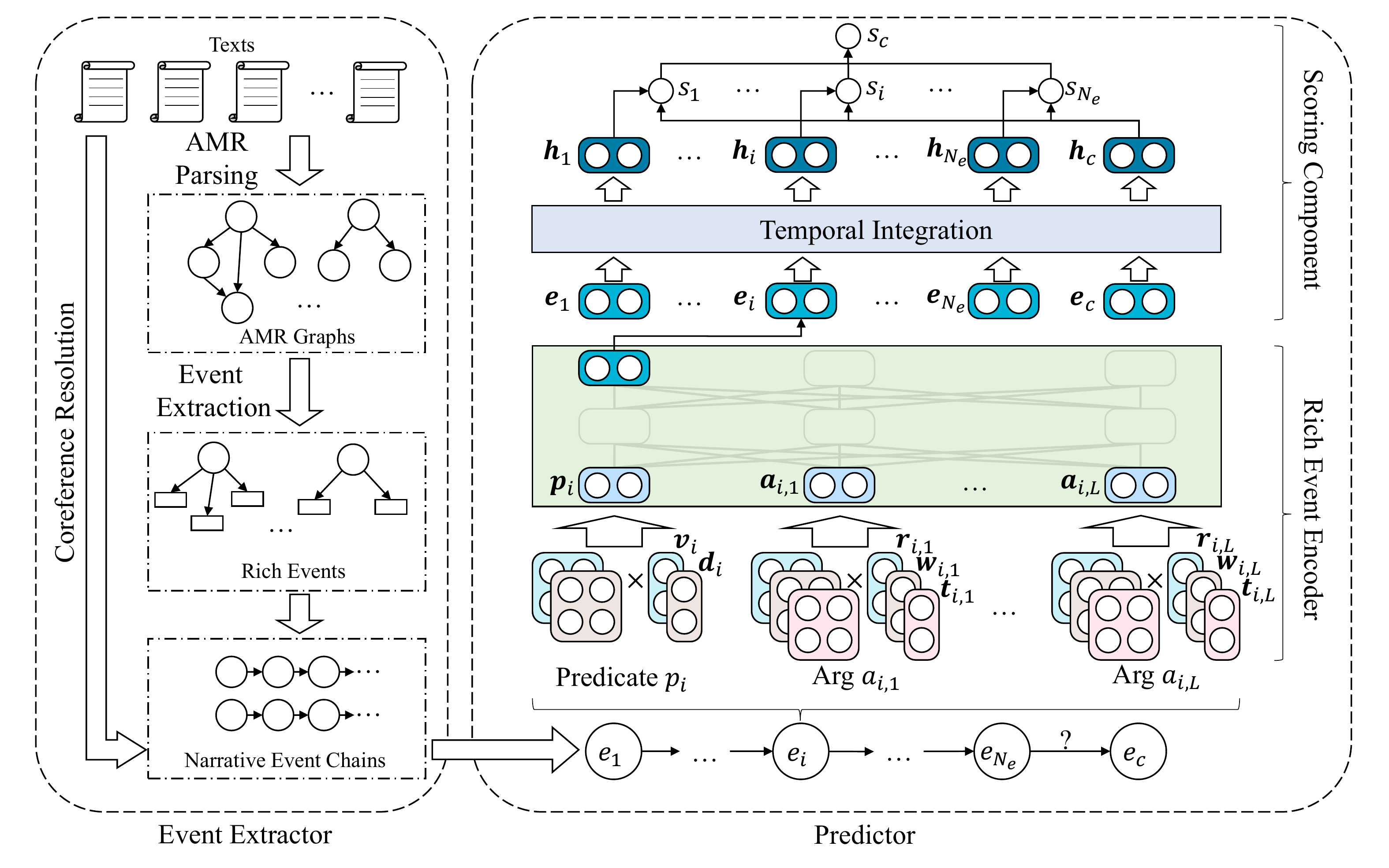}
	\caption{The overall architecture of REP.}
	\label{fig:model}
\end{figure*}

With respect to event encoding,
early studies apply one-hot encoding (i.e., symbolic event representation) to adapt to the counting-based scoring method, such as PMI~\cite{chambers-jurafsky-2008-unsupervised} or Bi-gram~\cite{jans-etal-2012-skip}.
As the number of arguments increases, this encoding method faces a severe sparsity problem.
Therefore, \citet{modi2014learning} embeds events into low-dimensional vectors via a shallow neural network.
Then, \citet{Granroth-Wilding_Clark_2016} apply an MLP to embed the events.
To capture subtle semantic interactions between the verb and each argument, \citet{Weber_Balasubramanian_Chambers_2018} adopt a tensor-based composition method.
It first maps each (verb, argument) pair into a vector,
and then aggregates these vectors to derive the event representations.
\citet{ding-etal-2019-event-representation} adopt a neural tensor network (NTN) based encoder to embed the events.
However, these methods are still not flexible enough to handle an arbitrary number of arguments.
In addition, they pay much attention to predicate-argument interactions
and underestimate the argument-argument interactions.

With respect to the relevance modeling,
early studies first compute the pairwise score between a candidate event and each event in the narrative event chain, and then aggregate these scores~\cite{chambers-jurafsky-2008-unsupervised,jans-etal-2012-skip,balasubramanian-etal-2013-generating,Granroth-Wilding_Clark_2016}.
These methods ignore the relevance between events in the narrative event chain.
Currently, there are two kinds of methods, namely, chain modeling and graph modeling.
Chain modeling methods view historical events as an event sequence~\cite{wang-etal-2017-integrating,Lv_Qian_Huang_Han_Hu_2019,bai-etal-2021-integrating},
while graph modeling methods view them as an event graph~\cite{ijcai2018-584,lee-goldwasser-2019-multi,lee-etal-2020-weakly}.
This paper does not aim to discuss this problem.
Therefore, we just adopt the widely-used chain modeling method.

\section{Preliminaries}
Currently, SEP follows the Multiple Choice Narrative Cloze (MCNC) setting, where the model is required to predict the most probable subsequent event $e^*$ from the candidate event set $\mathcal{C}$
according to the historical narrative event chain $\mathcal{H}$, i.e.,
\begin{equation}
	e^* = \arg \max_{e\in \mathcal{C}} \Pr(e | \mathcal{H}).
\end{equation}

Here, $\mathcal{H} = \{e_1, ..., e_{N_e}\}$ consists of $N_e$ historical events centered to the protagonist.
$\mathcal{C}=\{e_{c_1}, ..., e_{c_{N_c}}\}$ consists of $N_c$ candidate subsequent events.
Under our rich event description,
each event $e_i = (p_i, A_i)$ consists of the predicate $p_i=(v_i, d_i)$ and arguments $A_i=\{a_{i,j}\}$,
where $v_i$ is the sense of verb, 
$d_i$ is the semantic role of the protagonist, 
the $j$-th argument $a_{i,j} = (r_{i,j}, w_{i, j}, t_{i, j})$ consists of the semantic role $r_{i,j}$, the participant headword $w_{i,j}$, 
and the type $t_{i,j}$.

\section{Methodology}

In this section, we introduce the proposed REP framework.
As shown in Figure~\ref{fig:model},
it consists of two modules, namely the event extractor and the predictor.
% Firstly, we describe strategies to extract rich events and narrative event chains from texts.
% Then, we describe the predictor.
% Finally, we introduce the training details.

\subsection{The Event Extractor}

\begin{figure}
	\centering
	\includegraphics[scale=0.32]{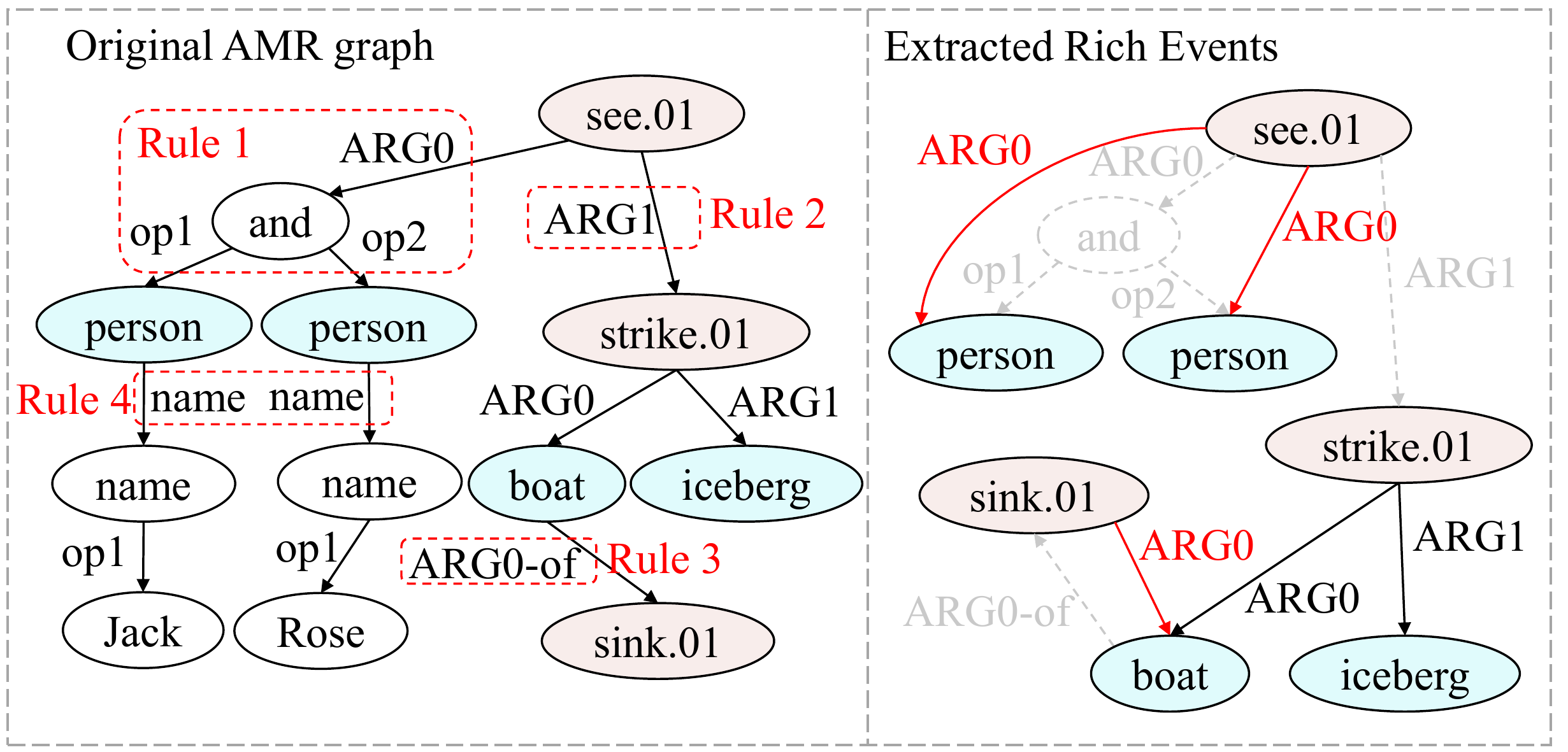}
	\caption{An example of applying rules to extract rich events (right) from AMR graph (left).
		The corresponding text is ``Jack and Rose see the boat striking an iceberg and sinking.'' 
		The verb sense nodes (pink) are seen as events, 
		and their children (blue) are seen as participants.}
	\label{fig:rules}
\end{figure}

Rich events require multiple kinds of information,
i.e., the senses of verbs, the semantic roles, the participants, and their types.
These kinds of information can be obtained via the AMR graphs of texts,
which consist of concepts (including senses of verbs and their participants) in texts and their semantic relations.
It is a unified semantic description framework that includes the above-mentioned information.
However, there exist some structural differences between the AMR graphs and the proposed rich event description.
Thus, we adopt the following rules to convert the AMR graphs into rich events, as shown in Figure~\ref{fig:rules},

\begin{itemize}
	\item \textit{Rule 1}: If a path follows 
	$X \xrightarrow{R} and \rightarrow Y $ pattern,
	we change it to
	$X \xrightarrow{R} Y$;
	
	\item \textit{Rule 2}: We remove all edges between two verb sense nodes; 
	
	\item \textit{Rule 3}: We change all $X \xrightarrow{\mbox{ARGN-of}} Y$ edges to
	$Y \xrightarrow{\mbox{ARGN}} X$;
	
	\item \textit{Rule 4}: We filter the edges according to their types.
	The reserved edge types are listed in Table~\ref{tab:rules},
	which mainly follow the definition in~\cite{zhang-ji-2021-abstract}.
	
\end{itemize}

\begin{table}
	\centering
	\begin{tabular}{lc}
		Categories & Edge Type                         \\ \hline
		core roles & ARG0, ..., ARG4                   \\
		operators  & op1, ..., op4                     \\
		spatial    & location, destination, path       \\
		means      & instrument, manner, topic, medium \\
		modifiers  & mod, poss, polarity               \\ \hline
	\end{tabular}
	\caption{Reserved edge types.}
	\label{tab:rules}
\end{table}

% Extract events
The verb sense nodes in the AMR graph are seen as the events,
and their children are seen as the types of participants.
% Connect events into chain
To obtain the participant headwords and construct the narrative event chains,
we align the participants in rich events to the coreferred entity mentions.
Specifically, we use a rule-based alignment tool~\footnote{RBW Aligner in https://github.com/bjascob/amrlib} to align AMR nodes to the tokens in texts.
Obviously, each participant corresponds to the root of a subtree in the original AMR graph.
The tokens that are aligned to the nodes in this subtree are seen as related tokens to the participant.
For each participant, if the rightmost token of all related tokens is in an entity mention (produced by coreference resolution),
we view this participant as identical to this entity.
According to the coreferred entities,
following the convention~\cite{chambers-jurafsky-2009-unsupervised},
we choose the most salient headword among all mentions.
Finally, we construct the narrative event chains via the temporal order of events~\footnote{Following the convention, we use the textual order of verbs to approximate the temporal order.}.

\subsection{The Predictor}

The predictor consists of two main components,
namely, the rich event encoder and the scoring component.

\subsubsection{Rich Event Encoder}

The rich event encoder aims to convert the rich events into machine-computable representations, i.e., low-dimensional vectors.
Firstly, it obtains the individual representations of the predicate and arguments via integrating the information in them.
For event $e_i$, it elements, i.e.,
$v_i$, $d_i$, $r_{i,j}$, $w_{i,j}$, and $t_{i,j}$, are all embedded into vectors of dimension $d_w$.
In what follows, the embedding of each element is represented by the same letter in boldface.
The representations of predicate $p_i$ and the arguments $a_{i,j}$ are calculated by,
\begin{eqnarray}
	\mathbf{p}_i &=& W_1^T \mathbf{v}_i + W_2^T \mathbf{d}_i + \mathbf{b},	\\
	\mathbf{a}_{i, j} &=& W_1^T \mathbf{r}_{i,j} + W_2^T \mathbf{w}_{i,j} + W_3^T \mathbf{t}_{i,j} + \mathbf{b},
\end{eqnarray}
where $W_1, W_2, W_3 \in \mathbb{R}^{d_w \times d_e}$ are projection matrices,
$\mathbf{b} \in \mathbb{R}^{d_e}$ is the bias vector.

After obtaining the predicate and argument representations,
the event encoder aggregates them to obtain the event representation.
To capture the subtle interactions between predicate and arguments,
previous methods~\cite{Weber_Balasubramanian_Chambers_2018,ding-etal-2019-event-representation} apply tensor-based networks to integrate each predicate-argument pair and then aggregate the results.
However, \citet{Weber_Balasubramanian_Chambers_2018} focus on the predicate-argument interactions and underestimate the argument-argument interactions.
\citet{ding-etal-2019-event-representation} adopt an additional tensor-based network to aggregate predicate-subject pair and predicate-object pair.
These manually designed steps that hierarchically aggregate predicate and arguments perform poor scalability when the number of arguments increases.
Another problem is that these methods often contain a huge number of parameters and thus require an expensive computational cost.
Though \citet{ding-etal-2019-event-representation} try to use low-rank tensor decomposition to decrease the number of parameters,
the number of parameters grows rapidly when the number of arguments increases,
which is far from solving this problem.

Considering the above problems,
we apply a multi-layer transformer network~\cite{NIPS2017_7181} to integrate the predicate and argument representations.
It has three advantages:
(1) It is able to sufficiently capture the interactions among the predicate and arguments;
(2) It is able to handle the increase of arguments with few human efforts, and thus, performs better in scalability;
(3) The number of parameters does not increase with the number of arguments.

The multi-head self-attention mechanism in this network is the key to enabling sufficient interactions among predicates and arguments.
Firstly, the predicate and argument representations are concatenated into a matrix $X=[\mathbf{p}_i, \mathbf{a}_{i,1}, ..., \mathbf{a}_{i, L}]$.
% X: (d_model, seq_len)
Then, $X$ is projected into the query matrix $Q=W_Q^T X$, the key matrix $K=W_K^T X$, and the value matrix $V=W_V^T X$,
where $W_Q, W_K, W_V \in \mathbb{R}^{d_e \times d_e}$ are projection matrices.
The three matrices are evenly split into $h_e$ slices.
The $k$-th head $head_k$ is calculated as follows,
\begin{equation}
	head_k = \mbox{softmax} (\frac{Q_k^T K_k}{\sqrt{d_e / h_e}})V_k,
\end{equation}
where $Q_k,K_k,V_k$ are the $k$-th slices of $Q,K,V$.

Finally, the heads are concatenated to compute the output,
\begin{equation}
	X' = [head_1^T, head_2^T, ..., head_h^T] W_O,
\end{equation}
where $W_O\in \mathbb{R}^{d_e \times d_e}$ is a projection matrix.
We use the output vector corresponding to the predicate as the event representation $\mathbf{e}_i$.

\subsubsection{Scoring Component}
After the historical events and the candidate events are encoded as $\mathbf{e}_i$ and $\mathbf{e}_c$ via the rich event encoder,
the scoring component aims to score each candidate event.
To integrate temporal order information into event representations,
we append each candidate event to the end of the historical narrative event chain~\cite{wang-etal-2017-integrating}.
Similar to \cite{bai-etal-2021-integrating},
we adopt a stacked transformer network with positional embeddings (Temporal Integration in Figure~\ref{fig:model}).
The outputs of the last transformer layer, $\mathbf{h}_i$ and $\mathbf{h}_c$,
are the temporal-aware event representations.

To obtain the advantage of event pair similarity and the temporal order information~\cite{wang-etal-2017-integrating},
we then calculate the pairwise score between each historical event and the candidate event,
\begin{equation}
	s_i = \mbox{sim}(\mathbf{e}_i, \mathbf{e}_c),
\end{equation}
where $\mbox{sim}$ is the negative Euclidean distance.

An attention weight is applied to each score to measure the importance of different event pairs,
\begin{equation}
	\alpha_i = \frac{\mathbf{e}_i^T \mathbf{e}_c}{\sqrt{d_e}}.
\end{equation}

Then, the score of the candidate event $e_c$ is calculated by adding up all weighted scores,
\begin{equation}
	s_c = \sum_{i=1}^{N_e} \alpha_i s_i.
\end{equation}

Finally, the probabilities of the candidate events are calculated by applying softmax on their scores,
\begin{equation}
	\Pr(e_{c_i} | \mathcal{H}) = \frac{\exp(s_{c_i})}{\sum_{j=1}^{N_c} \exp(s_{c_j})}.
\end{equation}

\subsubsection{Variants}
To verify the effectiveness of the proposed rich event encoder,
we also propose a simple fusion event encoder for rich events.
This simple encoder just aggregates the predicate and argument representations via adding them up, i.e.,
\begin{equation}
	\mathbf{e}_{i} = \sigma (\mathbf{p}_i + \sum_{j=1}^L \mathbf{a}_{i,j}),
\end{equation}
where $\sigma$ is a tanh activation function.
This encoder can be seen as an expansion of the current MLP event encoder~\cite{Granroth-Wilding_Clark_2016}.

\subsection{Training Details}
The training objective is to minimize the cross-entropy loss:
\begin{equation}
	L(\Theta) = -\frac{1}{N} \sum_{i}^{N} \log \Pr(e_i^* | \mathcal{H}_i) + \lambda || \Theta ||_2^2,
\end{equation}
where $e_i^*$ is the correct answer of the $i$-th sample;
$\mathcal{H}_i$ is the historical narrative event chain of the $i$-th sample; 
$N$ is the number of training samples;
$\Theta$ is the model parameters;
$\lambda$ is the L2 regularization factor.
The embeddings of verb sense $\mathbf{v}_i$, semantic role $\mathbf{r}_{i,j}$, headword $\mathbf{w}_{i,j}$ and type $\mathbf{t}_{i,j}$ are initialized randomly and trained together with other parameters.
The model is optimized by Adam~\cite{DBLP:journals/corr/KingmaB14} algorithm with 1000-size mini-batch.

\section{Experiments}

\begin{table}
	\centering
	\begin{tabular}{lcc}
		&   MCNC    & MCNC-rich \\ \hline
		\# Train Docs      &  830,645  &  75,466   \\
		\# Dev Docs        &  103,583  &   9,267   \\
		\# Test Docs       &  103,805  &   9,295   \\
		\# Train Instances & 1,440,295 & 1,006,301 \\
		\# Dev Instances   &  10,000   &  10,000   \\
		\# Test Instances  &  10,000   &  10,000   \\
		\# Arguments       &     3     &    23     \\
		Duration           & 1994-2004 & 1994-1996 \\ \hline
	\end{tabular}
	\caption{Dataset statistics on MCNC and MCNC-rich.}
	\label{tab:dataset}
\end{table}

\subsection{Datasets}
We use two datasets to evaluate the proposed framework.
Basic statistics of the two datasets are shown in Table~\ref{tab:dataset}.
\begin{itemize}
	\item \textbf{MCNC} dataset~\cite{Granroth-Wilding_Clark_2016} is extracted from 
	the New York Time portion of the Gigaword corpus~\cite{graff2003english}
	with events in the form of existing event description.
	Specifically, it uses news categorized as ``story'' from year 1994 to 2004.
	It utilizes the C\&C tool~\cite{curran-etal-2007-linguistically} for event extraction 
	and OpenNLP~\footnote{http://opennlp.apache.org} for coreference resolution.
	
	\item \textbf{MCNC-rich} dataset is proposed in this paper for the lack of rich event datasets.
	It is extracted from the same corpus with MCNC.
	Considering the computational cost, we only use news from year 1994 to 1996.
	We adopts SPRING parser~\cite{bevilacqua-etal-2021-one} for event extraction
	and AllenNLP~\cite{Gardner2017AllenNLP} for coreference resolution.
\end{itemize} 

Another widely-used dataset to evaluate the event modeling ability is the transitive sentence similarity dataset~\cite{DBLP:journals/corr/KartsaklisS14}.
However, this dataset represents events as $\langle subject, verb, object \rangle$ triples,
which is unsuitable for evaluating rich events.
Therefore, this dataset is not used.

\subsection{Experiment Settings}

The length of the narrative event chain $N_e$ is set to 8;
the number of the candidate events $N_c$ is set to 5;
word embedding dimension $d_w$ is set to 300;
event embedding dimension $d_e$ is set to 128;
the number of layers for rich event encoder is selected from $\{1, \underline{2}\}$;
the dimension of feedforward network in rich event encoder is selected from $\{512, \underline{1024}\}$;
the number of heads for rich event encoder is set to 8;
the number of layers for temporal integration is set to 2;
the dimension of feedforward network in temporal integration is set to 1024;
the number of heads for temporal integration is set to 16;
the dropout rate is set to 0.1;
the learning rate is set to 1e-3;
the regularization factor $\lambda$ is set to 1e-5;
The best settings (underlined) are selected according to the performance on development set.
All the experiments are conducted on Tesla V100.

\subsection{Baselines}
We apply the following representative methods as baselines:
1) \textbf{PMI}~\cite{chambers-jurafsky-2008-unsupervised} uses pointwise mutual information to measure the event pair similarity;
2) \textbf{Event-Comp}~\cite{Granroth-Wilding_Clark_2016} uses MLP to encode events and measure the event pair similarity;
3) \textbf{FEEL}~\cite{Lee_Goldwasser_2018} is an event modeling method, which considers the sentiments of events and the animacies of participants;
4) \textbf{SGNN}~\cite{ijcai2018-584} uses the narrative event evolution graph to describe the relevance among events and adopts a graph neural network to predict the subsequent event;
5) \textbf{SAM-Net}~\cite{Lv_Qian_Huang_Han_Hu_2019} combines a LSTM network with a DenseNet to encode the historical narrative event chain and predicts the subsequent event;
6) \textbf{SentInt}~\cite{ding-etal-2019-event-representation} is an event modeling method, which considers the sentiments and intentions of events;
7) \textbf{Lv2020}~\cite{lv-etal-2020-integrating} utilizes an external commonsense knowledge base;
8) \textbf{SCpredictor} and \textbf{MCPredictor}~\cite{bai-etal-2021-integrating} apply stacked transformer network to integrate temporal order information. MCPredictor utilizes multiple narrative event chain. We use the version that excludes text information.

For the MCNC-rich dataset,
considering the different structures between the rich events and existing events,
the baselines should be modified to adapt to this dataset.
Therefore, we compare REP with the baselines that only utilize information within a narrative event chain (i.e., PMI, Event-Comp, SAM-Net, and SCPredictor).
The other baselines use information out of a single narrative event chain,
such as other chains or external knowledge.
Thus, they are not directly applicable to this dataset.
For the MCNC dataset,
we compare REP with all the listed baselines.

\begin{table}
	\centering
	\begin{tabular}{lc}
		Method      & Accuracy (\%)  \\ \hline
		Random      &     20.00      \\
		PMI         &     31.44      \\
		Event-Comp  &     40.08      \\
		SAM-Net     &     51.50      \\
		SCPredictor &     54.33      \\ \hline
		REP(F,-RT)  &     55.97      \\
		REP(F,-T)   &     56.08      \\
		REP(F)      &     58.97      \\
		REP(-RT)    &     57.07      \\
		REP(-T)     &     57.30      \\
		REP(-S)     &     59.28      \\
		REP         & \textbf{60.08} \\ \hline
	\end{tabular}
	\caption{Experimental results on MCNC-rich dataset.}
	\label{tab:main_experiment}
\end{table}

\begin{table}
	\centering
	\begin{tabular}{lc}
		Method      & Accuracy (\%)  \\ \hline
		Random      &     20.00      \\
		PMI         &     30.52      \\
		Event-Comp  &     49.57      \\
		SAM-Net     &     54.48      \\
		SCPredictor &     58.28      \\ \hline
		FEEL        &     55.03      \\
		SGNN        &     52.45      \\
		SentInt     &     53.93      \\
		Lv2020      &     58.66      \\
		MCPredictor &     59.24      \\ \hline
		REP*        & \textbf{59.60} \\ \hline
	\end{tabular}
	\caption{Experimental results on MCNC dataset.
		Here, REP* adopts the existing event description.}
	\label{tab:subsidiary_experiment}
\end{table}

\subsection{Results on MCNC-rich}

\begin{figure}
	\centering
	\includegraphics[scale=0.25]{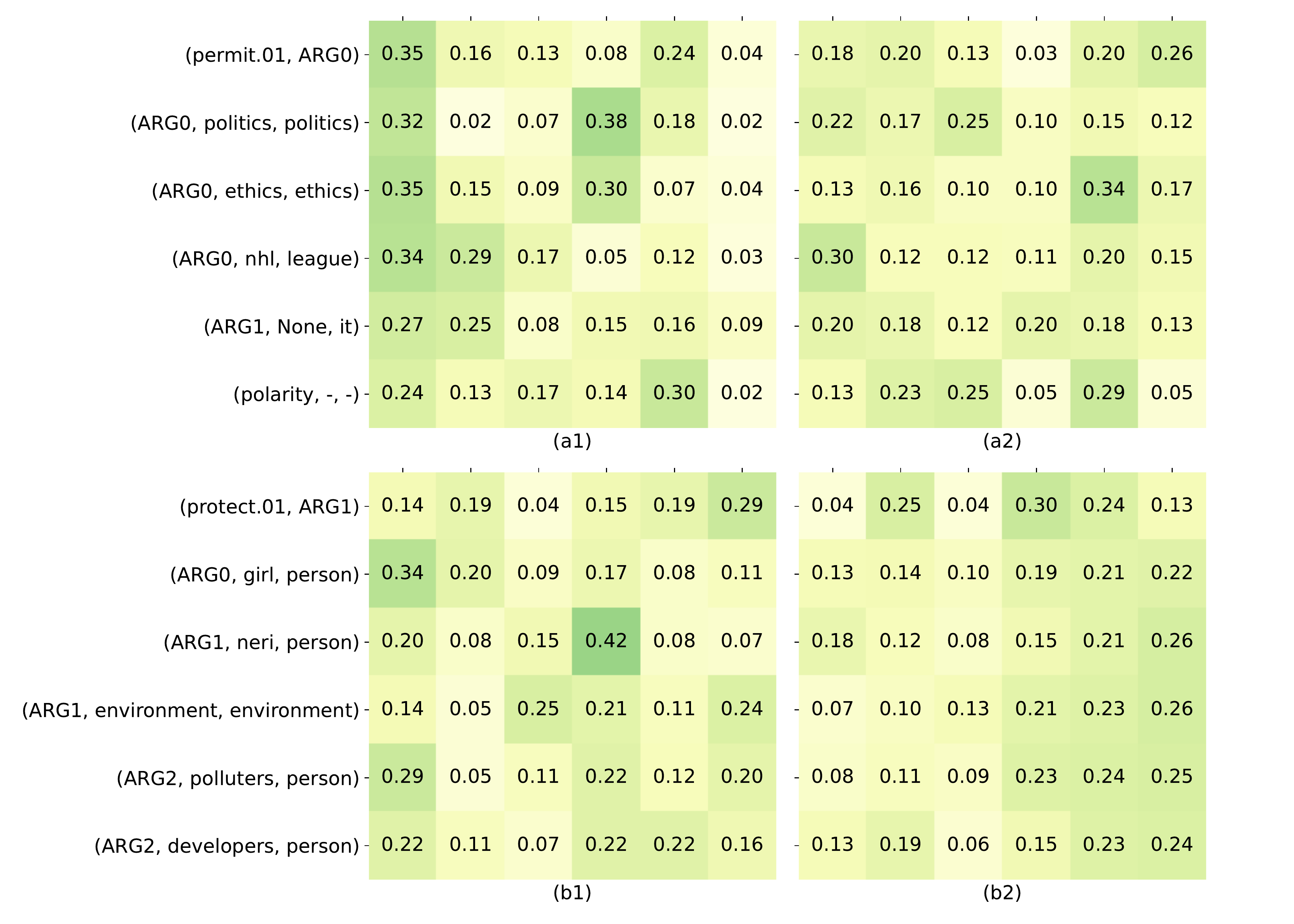}
	\caption{Attention heatmap for two events. 
		The text for (a1) and (a2) is ``The politics and very ethics of the NHL would never permit it .''
		The text for (b1) and (b2) is ``Ocean Girl is ... protecting Neri and the entire environment from corporate polluters and land developers .''}
	\label{fig:attn}
\end{figure}

% Basic settings
The experimental results on MCNC-rich dataset are shown in Table~\ref{tab:main_experiment}.
Here, ``F'' denotes the REP variant that applies the fusion event encoder;
``-S'' means that the model use verb lemmas instead of their senses;
``-T'' means that the types are not used;
``-RT'' means that both the extra semantic roles and the types are not used (i.e., only three kinds of semantic roles, namely, ARG0, ARG1, and ARG2, are considered).
From the results, we have the following observations:

\begin{figure*}
	\centering
	\includegraphics[scale=0.33]{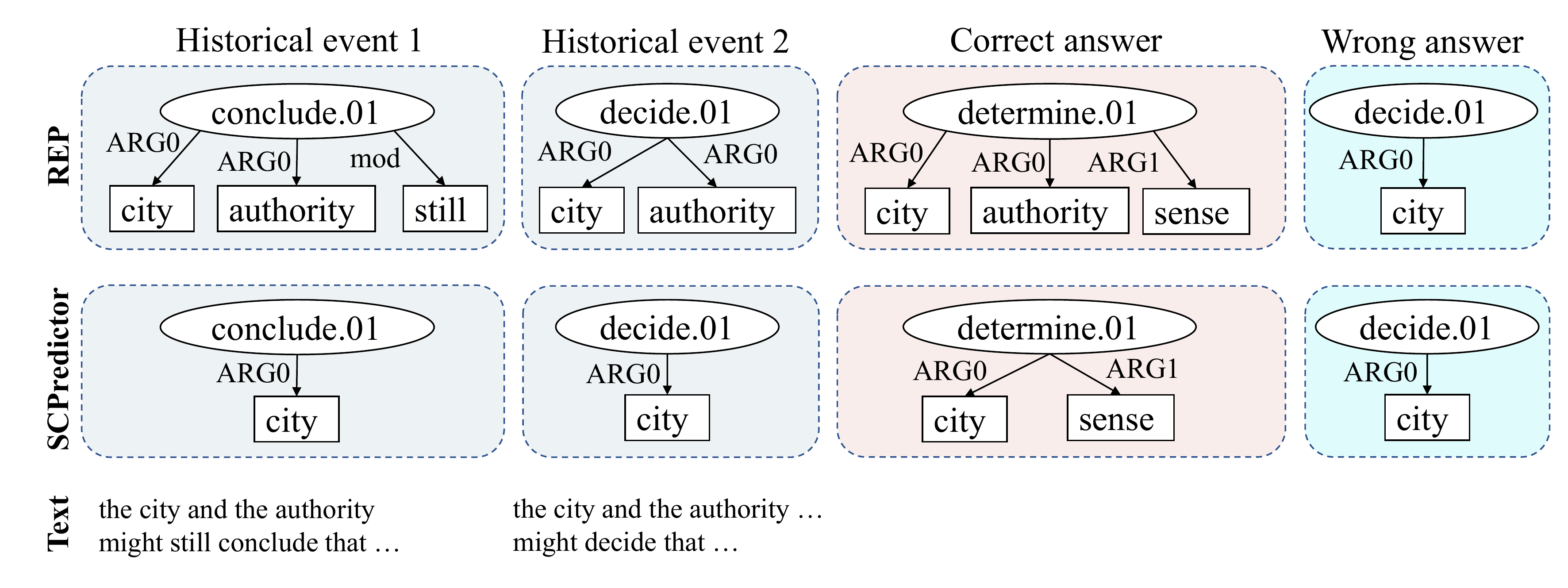}
	\caption{The case study on REP and SCPredictor, 
		where REP chooses the correct answer and SCPredictor chooses the wrong one. 
		We show the rich events used by REP and events used by the SCPredictor.
		The corresponding texts are also showed to help understanding the events.}
	\label{fig:case}
\end{figure*}

\begin{itemize}
	% 1. Compare the proposed model to other models
	\item REP outperforms the existing methods by more than 5.75\%,
	which shows that REP is able to effectively utilize the rich events,
	compared with the existing methods.
	
	% Add
	% 2. Ablation on verb sense
	\item REP outperforms REP(-S) by 0.80\%,
	which shows the importance of the verb senses.
	It is because multiple verb senses may refer to the same verb lemma,
	which brings difficulty in learning event representations.
	
	% 3. Ablation on semantic roles
	\item REP(F,-RT) outperforms SCPredictor by 1.64\%.
	Both methods use only ARG0, ARG1, and ARG2, 
	while REP(F,-RT) is able to handle multiple participants with the same semantic role (i.e., compound entities)
	so that the model is able to capture the co-occurrence among more participants.
	REP(F,-T) and REP(-T) outperform REP(F,-RT) and REP(-RT) by 0.11\% and 0.23\%, respectively,
	which shows that extra semantic roles
	%, such as location and modifiers, 
	are less important than those core semantic roles.
	
	% 4. Ablation on types
	\item REP(F) and REP outperform REP(F,-T) and REP(-T) by 2.89\% and 2.78\%, respectively,
	which shows the importance of types.
	It is because the types are more informative compared with the headwords.
	
	% 5. Ablation on event encoder (F, transformer)
	\item REP(-RT), REP(-T), and REP outperform REP(F,-RT), REP(F,-T), and REP(F) by 1.10\%, 1.22\%, and 1.11\%, respectively.
	These results show that the transformer-based rich event encoder is able to capture more subtle interactions among verb and arguments,
	compared with the fusion event encoder.
	
\end{itemize}

\subsection{Results on MCNC}

% Subsidiary experiment. MCNC
To further study the ability of the proposed rich event encoder under the existing event description (denoted as REP*),
we evaluate it on the MCNC dataset.
The experimental results are shown in Table~\ref{tab:subsidiary_experiment}.
Here, baselines are categorized into two parts.
The upper part (PMI, Event-Comp, SGNN, SAM-Net, and SCPredictor) consists of the baselines 
that only use the information within a narrative event chain.
In contrast, the lower part (FEEL, SentInt, Lv2020, and MCPredictor) consists of the baselines that use other information.
REP* outperforms all the baselines in the upper part by more than 1.32\% and outperforms the ones in the lower part by more than 0.36\%.
These results show that the proposed rich event encoder is able to capture more subtle interactions among the verb and the arguments than exiting event encoders,
even under the existing event description.
Especially, REP* does not utilize information out of a single narrative event chain,
and is still comparable to the newest baseline (MCPredictor), which uses other information.

\subsection{Analysis on Attention Weights}
% Heat map of event attention
% How different senses of a same verb affect attention?
% How the attention of the same senses in different chains look like?

To further study the interactions among the predicate and the arguments,
we study the self-attention heatmaps of two events in the development set of MCNC-rich,
where REP selects the correct answer, as shown in Figure~\ref{fig:attn}.
The attention weights are the average of all heads.
Figure~\ref{fig:attn} (a1) and (b1)  are the attention matrices from the first transformer layer,
while (a2) and (b2) are from the second layer.
Since REP uses the output corresponding to the predicate as the event representation,
in the second layer, only the weights corresponding to the predicate (row. 1) are involved in the calculation.
We have the following observations:

\begin{itemize}
	% Predicate-argument interaction
	\item In Figure~\ref{fig:attn} (a1) and (b1),
	the predicate usually has a relatively high weight when aggregating the representation of each argument (col. 1).
	This phenomenon is consistent with the conclusion of the previous studies that predicate-argument interactions are usually important~\cite{Weber_Balasubramanian_Chambers_2018}.
	
	% Argument-argument interaction
	\item Row. 2 col. 4 of Figure~\ref{fig:attn} (a1) and row. 3 col. 4 of Figure~\ref{fig:attn} (b1)
	both show that, in some cases, the argument-argument interactions are also important for modeling the events.
	These results also show that the proposed rich event encoder is able to consider these interactions.
	
	% Event aggregation
	\item In Figure~\ref{fig:attn} (a2) and (b2),
	we observe that the weights of arguments differ significantly (row. 1),
	which verifies the ability of REP to learn the impacts of arguments on the events.
	In addition, row. 1 col. 6 of Figure~\ref{fig:attn} (a2) shows that modifiers, such as negations,
	are able to play an important role when modeling the events.
	This result verifies our motivation to introduce extra semantic roles.
	
\end{itemize}

\subsection{Case Study}

To dive deep into the effects of REP,
we study the cases in the development set of MCNC-rich and compare with the best baseline SCPredictor.
As shown in Figure~\ref{fig:case},
the verb senses of the two candidates (determine.01 and decide.01) are similar.
Therefore, in this situation, the models should focus more on the arguments to derive the answer.
Both models obtain the same information from the wrong answer,
while REP obtains an additional participant ``authority'' from the correct answer.
According to history,
``authority'' and ``city'' frequently participate in the same events,
which implies that they are more likely to participate in the subsequent event together.
However, SCPredictor cannot handle multiple participants for the same semantic role.
Therefore, it is not able to capture such evidence and predicts the wrong answer.
This case shows the necessity to describe events by the proposed rich event description and the importance of argument-argument interactions.

\section{Conclusion and Future Work}

In this paper, we propose the REP framework for SEP.
To describe events more precisely, we propose the rich event description,
which enriches the existing ones with three kinds of important information,
namely, senses of verbs, extra semantic roles, and types of participants.
An event extractor is applied to extract rich events from texts.
To predict the subsequent event, the predictor adopts a rich event encoder that flexibly captures the subtle interactions among the verb and the arbitrary number of arguments.
Experimental results demonstrate its superiority.

However, we adopt a series of heuristic rules to convert AMR graphs and coreferred entities into rich events,
which still introduce noise.
It remains a challenge to obtain high-quality rich events.
In addition, when modeling participants, we only adopt headwords and types.
Other information, such as entity mention and the original text,
is not taken into consideration.
We will study these problems in the future.

\section*{Acknowledgements}
The work is supported by the National Natural Science Foundation of China under grants U1911401, 62002341 and 61772501, the GFKJ Innovation Program, Beijing Academy of Artificial Intelligence under grant BAAI2019ZD0306, and the Lenovo-CAS Joint Lab Youth Scientist Project.
Thanks to the reviewers for the constructive discussions and suggestions.

\bibliography{aaai23}

\end{document}